\documentclass{article}
\usepackage{times}
\usepackage{graphicx}
\usepackage{subfigure}
\usepackage{natbib}
\usepackage{algorithm}
\usepackage{algorithmic}
\usepackage{hyperref}
\usepackage{times}
\usepackage{epsfig}
\usepackage{stmaryrd}
\usepackage[mathscr]{eucal}
\usepackage{lineno}
\usepackage{color}
\usepackage{filecontents}
\usepackage{subfigure}
\usepackage{multirow}
\usepackage{verbatim} 
\usepackage{tabularx}
\usepackage{multirow}
\usepackage{textcomp,booktabs}
\usepackage{lineno}
\usepackage{stmaryrd}
\usepackage[mathscr]{eucal}
\usepackage{amsmath}
\usepackage{amssymb}

\def\q{{\bf q}}

\def\S{{\bf S}}
\def\x{{\bf x}}
\def\y{{\bf y}}

\def\0{{\bf 0}}
\def\1{{\bf 1}}

\def\RB{{\mathbb R}}

\def\eg{\emph{e.g. }}
\def\ie{\emph{i.e. }}
\def\argmax{\mathop{\rm argmax}}
\def\argmin{\mathop{\rm argmin}}

\def\etal{{\em et al.\/}\,}

\newtheorem{lemmas}{Lemma}

\newtheorem{propositions}{Proposition}

\graphicspath{{../figure/}}   

\usepackage[accepted]{icml2014}

\icmltitlerunning{ Collaborative Receptive Field Learning }

\begin{document}

\twocolumn[
\icmltitle{ Collaborative Receptive Field Learning }

\icmlauthor{Shu Kong}{aimerykong@gmail.com}
\icmladdress{HKUST, and Noah's Ark Lab of Huawei Co. Ltd.}
\icmlauthor{Zhuolin Jiang}{zhuolin.jiang@huawei.com}
\icmlauthor{Qiang Yang}{qiang.yang@huawei.com}
\icmladdress{Noah's Ark Lab of Huawei Co. Ltd.}

\icmlkeywords{Submodular Function, Image Matching, Translation Invariance, Scale Invariance, Object Categorization, Image Classification}

\vskip 0.3in
]

\begin{abstract}
The challenge of object categorization in images is largely due to arbitrary translations and scales of the foreground objects.
To attack this difficulty,
we propose a new approach called collaborative receptive field learning to extract specific receptive fields (RF's) or regions from multiple images,
and the selected RF's are supposed to focus on the foreground objects of a common category.
To this end,
we solve the problem by maximizing a submodular function over a similarity graph constructed by a pool of RF candidates.
However,
measuring pairwise distance of RF's for building the similarity graph is a nontrivial problem.
Hence, we introduce a similarity metric called pyramid-error distance (PED) to measure their pairwise distances through summing up pyramid-like matching errors over a set of low-level features.
Besides,
in consistent with the proposed PED,
we construct a simple nonparametric classifier for classification.
Experimental results show that our method effectively discovers the foreground objects in images,
and improves classification performance.

\end{abstract}

\section{Introduction}
\label{sec:intro}
It is widely known that
the difficulty in automatic object categorization from images is largely due to the arbitrary translations and scales of the foreground objects.
To solve the problem,
researchers have designed  robust image features like SIFT~\cite{lowe2004distinctive} and HoG~\cite{dalal2005histograms} for image representation,
reliable image (region) matching techniques such as image alignment~\cite{gunhee-cvpr-13} and detection~\cite{russakovsky2012object},
and sophisticated classifiers~\cite{duchenne2011graph}. 


Good image representation first concerns robust features.
Current feature learning methods propose to learn mid-level features hierarchically built over low-level ones,
which are also preferably learned adaptively rather than the hand-crafted ones~\cite{2013arXiv1311.2524G},
\eg SIFT and HoG.
Then,
various representation learning methods are proposed,
such as spatial pyramid~\cite{lazebnik2006beyond} and multiple layers of pooling and downsampling~\cite{krizhevsky2012imagenet}.
These representations can roughly preserve the salient object structures,
thus they enhance the discriminativeness of image representations.
With the adaptively learned features and discriminative representations,
the classification performance is improved accordingly.
But as these methods cannot effectively handle large translations and scales of the objects,
the accuracy gains are still limited.


There are some approaches attempting to localize the foreground objects for better encoding images,
such as saliency detection~\cite{van2011segmentation, fu2013cluster},
segmentation~\cite{chang2011co} and object detection~\cite{nguyen2009weakly, russakovsky2012object}.
Essentially,
these methods can be cast as the so-called receptive field learning as they intend to find the most desirable image regions (receptive fields) for particular tasks.
For example,
Jia \etal \ try to solve this problem by optimizing spatial pyramid matching (SPM) in building mid-level features~\yrcite{jia2012beyond}.
Their method selectively combines pooled features in predefined image regions to improve the discriminability of overall image representations.
However,
Their method using mid-level features learns the same combination patterns for all images of different categories,
thus it still fails in handling prominent translations and scales in individual images.

To effectively handle arbitrary scales and translations of objects,
we propose a new framework called \textbf{collaborative receptive fields learning} (coRFL),
which intends to discover specific receptive fields (RF's) or image regions that mainly cover the foreground objects from the same category.
coRFL merely requires weak labels~\cite{russakovsky2012object};
that is to say,
there is no exact object location information but the category-level label for each image.
Moreover,
coRFL learns to find these RF's collaboratively among multiple images from a common category,
thus leading to reciprocal accuracies for discovering their common object.
Note that some definitions of receptive field in neural science are different to ours~\cite{olshausen1996emergence},
but we keep using this term to highlight the meaning that RF's in images received
by the computer should capture the most distinct foreground object.
We model coRFL as selecting specific vertices from a graph,
which is constructed by pairwise similarities of RF candidates from these images.
Borrowing some vision-based priors,
we formalize the problem as a submodular function,
with which a simple greedy method suffices to produce performance-guaranteed solutions.

However,
in building the graph,
finding the right metric of pairwise distance between RF's with varying sizes is a nontrivial problem.
One intuitive way to represent the RF is to use the mid-level feature concatenated by multi-layer pooled vectors~\cite{yang2009linear} with the same length.
But these features usually have thousands of dimensions,
and will lose distinct information due to its vector quantization or sparse coding~\cite{DomingosCommACM, boiman2008defense}.
For this reason,
we introduce Pyramid-Error Distance (PED),
a nonparametric method to measure the distance of image regions over sets of low-level SIFT features.

We perform coRFL in training images of each category to purify the training set by only preserving RF's that capture the meaningful foreground objects.
With the proposed PED,
we design a nonparametric classifier to match qeury images with the purified training set.
Through experiments over both synthetic data and benchmark databases,
we show the effectiveness of our proposed framework.

\textbf{Contributions and Paper Organization: }
We first review essential preliminaries in Section~\ref{sec:relatedWork}.
Then we elaborate our framework of coRFL in Section~\ref{sec:cRFLearning},
the metric of Pyramid-Error Distance (PED) in Section~\ref{sec:PED},
and our designed nonparametric classifier in Section~\ref{sec:classifier},
respectively.
We evaluate our framework with experiments in Section~\ref{sec:exp},
before concluding in Section~\ref{sec:conclusion}.

\section{Related Work}
\label{sec:relatedWork}

There are three keywords in our framework,
receptive field learning,
submodular function and similarity metric.

\textbf{Receptive Field Learning:}
Multiple problems in computer vision can be seen as receptive field learning to aid image understanding.
For example,
saliency detection aims to discover regions that capture human attention in the images with perceptual biases;
the result of detected salient regions anticipate better image matching by only considering these salient regions~\cite{fu2013cluster}.
Besides,
image segmentation aim to simplify or change the representation of an image into something that is more meaningful and easier to analyze~\cite{shi2000normalized}.
Requiring the localization information,
van de Sande \etal \ use salient regions through multi-scale segmentation with multiple cues to detect the object of interest~\yrcite{van2011segmentation}.
Moreover,
by considering object translations and scales,
Russakovsky \etal \ propose object-centric spatial pooling (OCP) approach~\yrcite{russakovsky2012object},
which first infers the location of the objects
and then uses their locations to pool foreground and background regions separately to form the mid-level features.
OCP learns the object detectors with the weak labels,
\ie there is no exact object location information in images.
This is the same condition in our work.


In particular,
Jia \etal \  explicitly work on receptive field learning through learning to selectively combine pooled vectors over 100 predefined grids~\yrcite{jia2012beyond},
as demonstrated by Fig.~\ref{fig:RFtemplatesComparison} (b).
But their method learns the same combination pattern across all images from different categories.
Therefore,
when facing notable translation and scale changes of foreground objects,
it cannot be guaranteed to achieve improved performance.
Moreover,
Duchenne \etal \ introduce a graph-matching kernel (GMK) to address object deformations
between every pair of images~\yrcite{duchenne2011graph}.
However,
the kernel requires more time to calibrate images for large-scale datasets,
and also fails in calibrating images with extremely cluttered backgrounds

\textbf{Submodular Function:}
The natural and wide applicability of submodular function makes it receiving more attention in recent years~\cite{Iyer2013icml}.
Let $\cal V$ be a finite ground set.
A set function ${\cal F}: 2^{\cal V} \rightarrow \RB$ is submodular if ${\cal F}(A\cup a) - {\cal F}(A) \ge {\cal F}(A\cup \{a, b \}) - {\cal F}(A\cup b) $,
for all $A \subseteq {\cal V}$ and $a, b \in \bar A$.
Here,
$\bar A = {\cal V}/A$ is the complement of $A$.
The property is referred to as diminishing return property,
stating that adding an element to a smaller set helps more than adding it to a larger one.
As for other properties of submodularity,
please refer to~\cite{Iyer2013icml} and references therein.


\textbf{Similarity Metric:}
The similarity measurement of image regions is still an open problem.
Recently,
the combination of mid-level features and SVM classifier generally produce promising results~\cite{lee2009convolutional, boureau2010learning, zeiler2011adaptive}.
The mid-level features are usually generated by concatenating multi-layer (pooled) features within spatial pyramid pattern as demonstrated in Fig.~\ref{fig:RFtemplatesComparison} (a)~\cite{lazebnik2006beyond, yang2009linear, coates2011importance},
or learned through the convolutional neural networks (CNN)~\cite{krizhevsky2012imagenet, 2013arXiv1311.2524G}.
However,
pairwise similarity between mid-level features cannot be reliably measured by their Euclidean distance,
due to both their high dimension and the vector quantization or sparse coding stage in extracting mid-level features~\cite{boiman2008defense, DomingosCommACM}.
Among these methods,
the Euclidean distance over low-level SIFT descriptors~\cite{boiman2008defense} motivates our proposed metric.

\section{Collaborative Receptive Field Learning}
\label{sec:cRFLearning}
In this section,
we present the proposed framework of collaborative receptive field learning (coRFL) in detail.
With the weak labels and fed multiple images of a common category,
coRFL collaboratively extracts specific receptive field (RF's) that capture the common foreground objects.
Solving the problem is the core in our proposed framework,
because we perform coRFL over training images in each category to purify the training set for matching queries,
and the resultant images only preserve the most meaningful foreground objects.
We first demonstrate how to extract RF candidates,
and then present some vision-based priors before the formalism of coRFL.


\subsection{Extracting Receptive Field Candidates}

\begin{figure}[t]
        \centering
        \includegraphics[width=.99\linewidth]{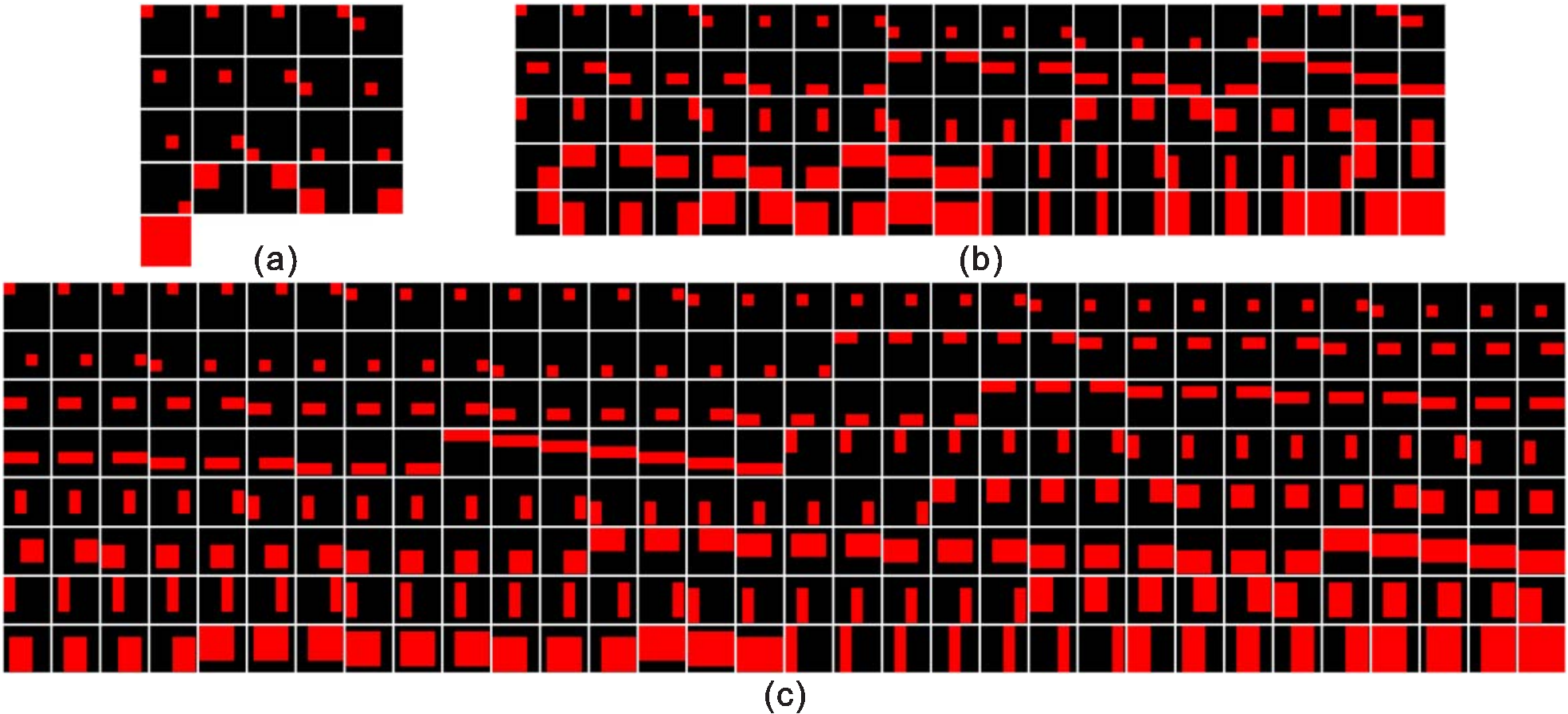}
    \caption{ Comparisons of receptive field (RF) candidates:
    (a) spatial pyramid grids~\cite{yang2009linear};
(b) RF's for selective combination~\cite{jia2012beyond};
(c) RF candidates in our method. }
    \label{fig:RFtemplatesComparison}
\end{figure}

Suppose there are $N$ images available from a specific category,
without loss of generality,
we predefine $m$ templates to extract RF candidates in images.
In this work,
we define $m=256$ candidates as shown in Fig.~\ref{fig:RFtemplatesComparison} (c),
leading to $M=mN$ candidates in total for these $N$ images.
In contrast to the approach that defines 100 grids~\cite{jia2012beyond},
our overlapping grids can capture the foreground objects more reliably and correctly,
thus preventing the computation from covering only object parts or too many image backgrounds.
Now,
we solve coRFL by selecting the most desirable image regions or receptive fields (denoted by $RF_+$) that  capture the common and distinct objects from these images.
In particular,
we can specify the number of selected $RF_+$'s as $K$.
Then,
the crucial point is to sketch a mechanism to find the most desirable $RF_+$'s that are distinct from the negative ones (denoted by $RF_-$ that mainly covers the cluttered background).

It is worth noting the difference between our RF candidates and the pooled features of image grids~\cite{lazebnik2006beyond}.
Specifically,
our method finds the most desired RF's that capture the foreground objects from predefined grids;
and these selected RF's form the training set used for classification.
In contrast,
SPM-based methods (selectively) concatenate the pooled vectors for the final representation of the overall image,
such as in~\cite{yang2009linear, jia2012beyond}.
As a result,
only our method explicitly considers object scale and translation in individual images.

In addition,
Girshick \etal \ propose to extract region proposals (with different sizes) in images for object detection~\yrcite{2013arXiv1311.2524G}.
These region proposals can also be seen as RF candidates,
but they are required to be warped with brutal force into a fixed size,
so that they can be fed into a CNN~\cite{krizhevsky2012imagenet}.
Different from these methods,
this transformation may destroy information related to object appearance and shape.
In contrast,
our method preserves such valuable information by allowing various sizes of the RF's.

\subsection{Inter- and Intra-Image Prior }

Inspired by the area of saliency detection,
we assume that the $RF_+$ capturing the object is more salient than others ($RF_-$'s),
which mainly cover the background or object parts.
In other words,
there should be a large contrast between $RF_+$ and $RF_-$,
both from the image itself and other images.
We call  intra- and inter-image prior, respectively.
Therefore,
an oracle should find (a few) $RF_+$'s which have small similarities with (most) $RF_-$'s,
\ie
pairwise similarities between selected $RF_+$ and $RF_-$ should be small.

Besides,
as multiple images are given from a common category,
we can say each image has at least one $RF_+$ that makes them correlated semantically by capturing the common objects.
Based on this repeatedness principle,
inter-image relationship can be exploited by considering similarities between $RF_+$'s from different images should be large.
Moreover,
pairwise similarities between $RF_-$'s should be small,
since $RF_-$'s mainly capture cluttered background which can be seen as noises.

To model the priors,
we build a graph $\S \in \RB^{M \times M}$ to record the similarities between each pair of RF candidates.
The larger element, say $s_{ij}$ in $\S$,
means that receptive fields $i$ and $j$ are more similar to each other.
As the measurement is a nontrivial problem over RF's,
we continue to elaborate coRFL and put graph construction in Section~\ref{sec:PED}.
Now,
an oracle will find a set of vertices as the $RF_+$'s indexed by $A$,
such that the sum of similarities within $RF_+$'s is a maxima.
Meanwhile,
the summed similarity between $RF_+$'s and $RF_-$'s indexed by the complement $\bar A$,
as well as summed similarity within $RF_-$'s,
is a minima.


We define the following operation over matrix $\S$ with two sets $A$ and $B$ indexing rows and columns,
respectively:
\begin{equation}\small
\S_{A,B} = [s_{ij}] \in\RB^{\vert A \vert \times \vert B \vert}, \forall i\in A, j \in B.
\end{equation}
Note that $\S_{A,B}  = \S_{B, A}^T$ as we require the symmetric matrix $\S $ to specify an undirected graph.
Moreover,
over $\S_{A,B} \in \RB^{\vert A\vert \times \vert B \vert}$,
we define a function $h(\S_{A,B})$ for the sum of pairwise similarities between $A$ and $B$ as:
\begin{equation}\small
h(\S_{A,B}) =
\begin{cases}
\sum_{i\in A} \sum_{j\in B} c_{ij}, & A\not=\varnothing \text{ and }  B\not=\varnothing, \\
0, & \text{otherwise.}
\end{cases}
\end{equation}

Therefore,
we can say the $RF_+$'s indexed by  $A$ should simultaneously lead to a maxima of $h(\S_{A, A}) $,
a minima of $h(\S_{A, \bar A}) $ and $h(\S_{\bar A, \bar A}) $.
By unifying all these terms,
an oracle should find $A$ that leads to the maxima of the following:
\begin{equation}\small
{\cal H} (A) = \log \big( \mu + h(\S_{A,A}) - \alpha h(\S_{A,\bar A}) - \beta h(\S_{\bar A, \bar A}) \big),
\label{eq:similarityTerm}
\end{equation}
in which $\mu $ is a constant scalar\footnote{Generally, we set $\mu = 1+\tau h(\S_{\cal V, \cal V})$, where $\tau$ is defined as in Lemma~\ref{lemma:tau}.} that is sufficient large to ensure $ \mu + h(\S_{A,A}) - \alpha h(\S_{A,\bar A}) - \beta h(\S_{\bar A, \bar A})$ is positive;
and positive parameters $\alpha$ and $ \beta$ jointly control relative importance of the terms.
The $\log$ operation in Eq.~\ref{eq:similarityTerm} is used to ensure the following submodular property\footnote{All proofs to propositions and lemmas are presented in the supplementary material.}:
\begin{propositions}
There exist proper $\alpha$ and $ \beta$ that make the proposed function ${\cal H}(A)$ in Eq.~\ref{eq:similarityTerm}: $2^{\cal V} \rightarrow \RB$
a monotonically increasing and submodular function.
\label{proposition:Similarity}
\end{propositions}
Especially,
we have:
\begin{lemmas}
For any $\tau \in \RB^{*}$,
${\cal H}(A)$ in Eq.~\ref{eq:similarityTerm} is a monotonically increasing and submodular function
by setting $\alpha =\tau-1$ and $\beta= \tau$.
\label{lemma:tau}
\end{lemmas}
Following Lemma~\ref{lemma:tau},
we set $\tau > 1$ and require $\beta = \tau$ and $\alpha = \tau-1$
to benefit from the desirable submodularity and monotonicity of ${\cal H}(A)$,
and to model the principles in finding $RF_+$'s.
Hereafter, we rewrite as ${\cal H}_{\tau} (A)$ to explicitly highlight the sole parameter $\tau$.

\subsection{Balance Penalty}
Since the common weak labels enable images from one category to be correlated to each other,
it is reasonable for each image to contribute one $RF_+$ by itself.
Thus,
we should extract at least one $RF_{+}$ from each training image.
A benefit of this balance is the preservation of intra-class variability,
which helps alleviate the overfitting problem.

We call this balance principle,
\ie the number of $RF_{+}$ in each image should be balanced.
Specifically,
let $A_j$ index the $RF_+$'s from the $j^{th}$ image,
and $A=\cup_{j=1}^{N} A_j$ indices all the positive $RF_+$'s.
We add a penalty term to our objective function as below to balance the number of $RF_+$'s in the images:
\begin{equation}\small
\begin{split}
{\cal G}(A) = & \sum_{j=1}^N \log( \vert A_{j}\vert + 1), \\
\text{s.t. }  \cup_{j=1}^{N}& A_j = A,  A_i\cap A_j = \varnothing, \forall i\not=j,
\end{split}
\label{eq:numberBalance}
\end{equation}
where $\vert A_j \vert$ means the cardinality of index set $A_j$.
Particularly,
we have $\vert \varnothing \vert = 0$.
To understand the functionality of Eq.~\ref{eq:numberBalance},
please consider the following proposition:
\begin{propositions}
With the function defined as below over vector $\x=[x_1,\dots,x_N] \succcurlyeq 0$:
\begin{center}
$ g(\x) = \sum_{j=1}^N \log( x_j + 1), $ \\
\end{center}
if $x_c \le x_i, \forall i \not = c$,
then $g(x_c+1 \vert \x) \ge  g(x_i+1 \vert \x)$.
where $g(x_i+1 \vert \x) = \sum_{ j\not=i}  \log( x_j + 1) + \log( x_i + 1 + 1).$
\label{proposition:property_balance}
\end{propositions}
This property demonstrates that adding smaller elements achieves greater reward.
In particular,
over the graph defined by $\S$,
the vertices (RF's) are preferred to be selected from images one after another.
As a result,
balancing the number of elements can be achieved.

Furthermore,
we have the following proposition:
\begin{propositions}
The function ${\cal G}(A)$ in Eq.~\ref{eq:numberBalance}: $2^{\cal V} \rightarrow \RB$ is a monotonically increasing and submodular function.
\label{proposition:balance}
\end{propositions}

\subsection{Center-Bias Principle}
Inspired by the saliency detection,
we exploit center-bias principle~\cite{tatler2007central} to mildly constrain the $RF_+$ w.r.t its location in the image.
Specifically,
an $RF_+$ appears around the center of the image with high probability.
Our mild center-bias constraint means that,
searching for $RF_+$'s should focus more around the image center,
but still allows to capture the most desired  $RF_+$ locating near the image margin with high fidelity.

Intuitively,
center bias can be modeled through the position of RF's.
Let  ${\boldsymbol \theta} \in\RB^{M}$ denote the distances\footnote{Here the distance does not necessarily mean the Euclidean distance. For proper constraint, Euclidean distance with a Gaussian kernel is preferred in our work.}
of all the RF's to the image center,
specifying center-bias constraint for each of the $M$ RF candidates.
One intuitive example is to constrain that $\Vert {\boldsymbol \theta}_A \Vert_1$ to be small,
where ${\boldsymbol \theta}_A$ means a subvector comprising of elements indexed by $A$.
We specify $\Vert \q_A \Vert_1 = 0$ for $A = \varnothing$.
Alternatively,
if we store the reciprocal of center distances in $\q = [ q_k ] = [\frac{1}{\theta_k}], k=1,\dots,M$,
then we need to constrain that $A$ leads to a relatively larger $\Vert \q_A \Vert_1$,
maximizing which pushes our mechanism to focus on RF's around the images' centers.

\subsection{Objective Function} 

With the proposed inter- and intra-image prior, 
the balance penalty ${\cal G}(A)$, 
and the mild center-bias penalty, 
we turn to maximize the following objective function to find $RF_+$'s indexed by $A$:
\begin{equation} \small
\begin{split}
A = &\argmax_{ A \in {\cal I}}  \big\{ {\cal F}(A) \equiv {\cal H}_{\tau} (A) +  \lambda_1 {\cal G}(A) + \lambda_2 \Vert \q_{A}\Vert_1 \big\}\\
&\text{s.t. } \vert A \vert \le K,
\end{split}
\label{eq:obj}
\end{equation}
where $\lambda_1$ and $\lambda_2$ are the parameters to control
inter- and intra-image prior term, the balance penalty term and center prior term,
respectively.
With the proposition as below,
we can see that exactly $K$ receptive fields are extracted.

\begin{propositions}
            The function $ {\cal F}(A)$ in Eq.~\ref{eq:obj}: $2^{\cal V} \rightarrow \RB$ is a monotonically increasing and submodular function,
            and induces a uniform matroid ${\cal M} = ({\cal V}, {\cal I})$,
            where ${\cal V}$ is the point set, and $\cal I$ is the collection of subsets $A \subseteq {\cal V}$.
\label{proposition:obj}
\end{propositions}

Submodularity described by the above proposition indicates that a simple greedy algorithm suffices to produce performance-guaranteed solutions with a theoretical approximation $(1-1/e)$~\cite{nemhauser1978analysis}.
The greedy search requires $\vert {\cal V} - A \vert$ evaluations for the marginal gains before adding a new element into $A$ at each iteration.
To speed up the optimization process,
we use the lazy greedy~\cite{leskovec2007cost} by constructing a heap structure over marginal gains of each element.
Even through the addition of any element into $A$ impacts the gains of the remaining ones,
we can merely update the gain of the top element in the heap,
instead of recomputing the gains for every remaining element.
The key idea is that the gain for each element can never increase due to the diminishing return property of submodular function, which can be illustrated by the naive search method in Fig.~\ref{fig:syntheticDemonstration}.
Moreover,
the recomputation of gain for the top element in the heap is not much smaller in many cases,
hence the top element will stay the top element even after the update.

The worst case is to update the gain for each element and then re-establish the heap after the addition of any elements to $A$,
leading to the complexity $O(\vert {\cal V}\vert \log \vert {\cal V}\vert)$ for rebuilding the heap,
and the overall complexity $O(\vert {\cal V}\vert^2 \log \vert {\cal V}\vert)$~\cite{Cormen2001introduction} of the optimization.
But in practice,
the lazy algorithm only requires a few updates in the heap at each iteration.
Hence,
the complexity of the optimization is effectively $O(\vert {\cal V}\vert \log \vert {\cal V}\vert)$.

\section{  Similarity Graph Construction via Pyramid-Error Distance} %
\label{sec:PED}
In this section,
we investigate how to measure the pairwise similarity of RF's in constructing the graph.
As the RF's are essentially image regions with varying sizes,
measuring them is a nontrivial problem.
One intuitive idea is to borrow the mid-level pooled features~\cite{jia2012beyond} to represent each RF candidates,
as the pooling process generates feature vectors with fixed length.
But it will produce disastrous results due to both high dimensionality and vector quantization or sparse coding.
Therefore,
we introduce a new metric called Pyramid-Error Distance (PED).

\begin{figure}[t]
        \centering
        \includegraphics[width=.99\linewidth]{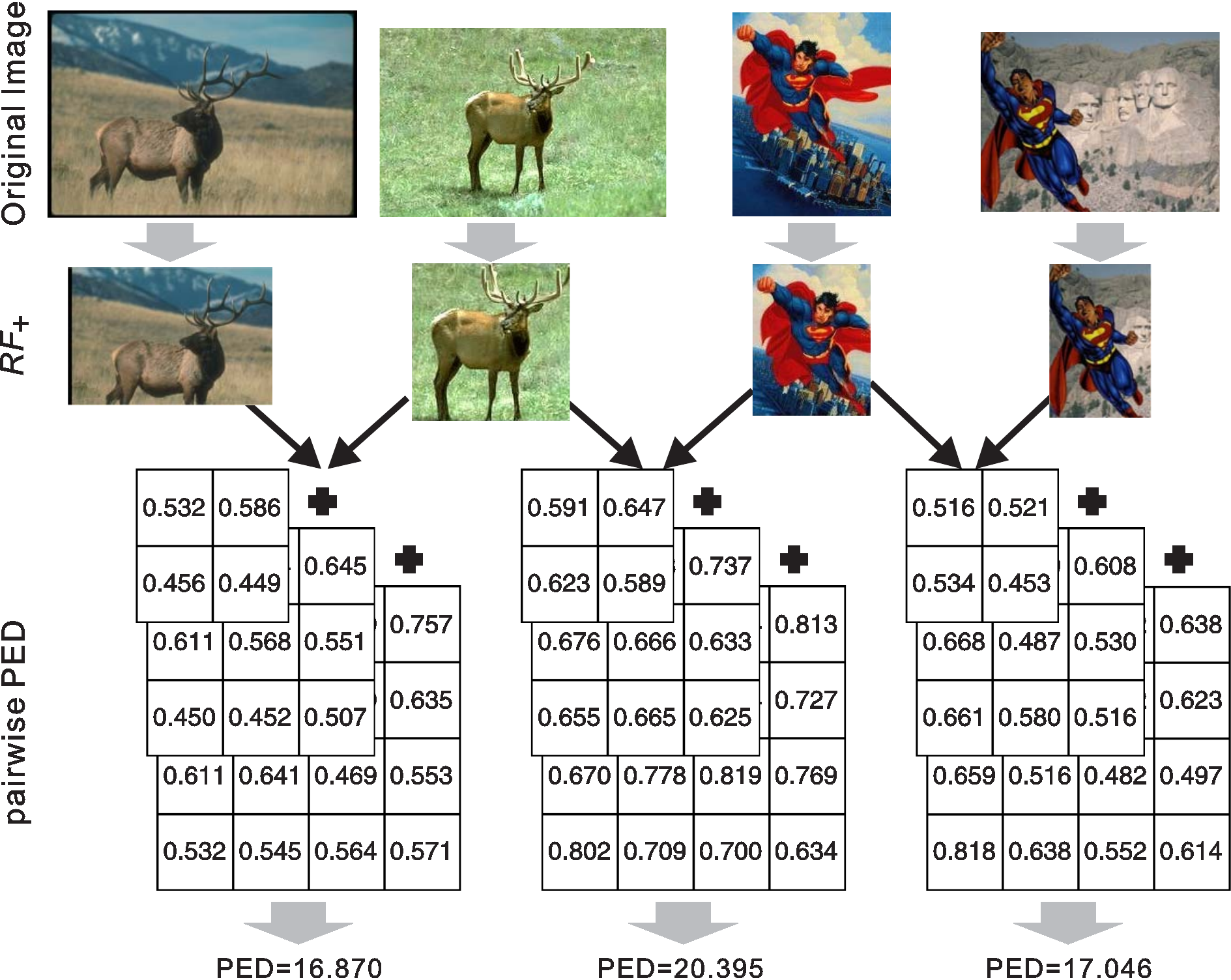}
    \caption{
     Illustration of Pyramid-Error Distance (PED).
    The first row presents the original images,
    and their corresponding RF's are learned by our algorithm in the second row.
    The third row shows the pairwise PED.
    PED encourages the similarity between RF's from the same class is larger than that from different classes.}
    \label{fig:PEDdemonstration}
\end{figure}

\subsection{Pyramid-Error Distance}
We split each  RF candidate into pre-defined grids at multiple levels.
For instance,
we use three partition scale $2\times 2$, $3\times 3$ and $4\times 4$,
leading to $L=29$ grids in total.
Please note that the pyramid partitions is done in each single RF,
instead of the overall image.
This is totally different from SPM-based methods~\cite{lazebnik2006beyond, boureau2010learning, jia2012beyond} that concatenate the pooled vectors of all grids to represent the whole.

Throughout our work,
we extracted the low-level SIFT features over each grid of  RF.
Before measuring the similarity of two RF's,
we first calculate the distance of a pair of grids from two RF's at a corresponding position indexed by $l$.
Let $X_l = \{\x_i \in \RB^p \vert i=1,\dots, r \}$ and $Y_l =\{\y_j \in \RB^p \vert j = 1,\dots, q\}$ be two sets consisting of $p$-dimensional descriptors,
representing the two corresponding grids, respectively.
As a result,
even with various sizes, grids can be represented by sets of low-level features.
Please also note the descriptor number $r$ and $q$ are not necessarily equal,
due to the number of feature points automatically detected in the RF's of different sizes.
We define the distance at set level as below:
\begin{equation} \small
\begin{split}
dist(X_l \vert\vert Y_l) =& \frac{1}{2r}\sum_{i=1}^{r} \big( \min_{j} \Vert \x_i - \y_j \Vert_F^2 \big)+ \\
& \frac{1}{2q}\sum_{j=1}^{q} \big( \min_{i} \Vert \x_i - \y_j \Vert_F^2 \big).
\end{split}
\label{eq:SetDist}
\end{equation}


Furthermore,
let $RF_i = \{X_l \vert l=1,\dots,L \}$ and $RF_j = \{Y_l \vert l=1,\dots,L\}$ be two RF's,
as shown by the second row in Fig.~\ref{fig:PEDdemonstration}.
Then, with the pyramid partitions,
we now arrive at the PED between two RF's as:
\begin{equation} \small
\begin{split}
D(RF_i \vert\vert RF_j) = \sum_{l} dist(X_l \vert\vert Y_l), 
\end{split}
\label{eq:SetDistPED}
\end{equation}
in which $l$ indexes the grid in specific location within a defined pyramid partition.
From the third row in Fig.~\ref{fig:PEDdemonstration},
the PED is calculated by the sum of grid distances in a pair of RF's.
To  analyze the complexity,
we naively assume there are $n_{grid}$ descriptors (with $d$-dimensionality) in each grid,
then calculating PED is of complexity $O(dLn^2_{grid})$.

\subsection{ Similarity Graph Construction }
\label{ssec:similaritygraphconstruction}
With the defined PED in Eq.~\ref{eq:SetDistPED},
we calculate the similarities of each pair of  RF's, and construct the graph $\S$ accordingly.
In detail,
over two receptive fields $RF_i$ and $RF_j$,
we calculate their PED as $D(RF_i \vert\vert RF_j)$,
and then transform PED into similarity $s_{ij}$ by a Gaussian kernel:
\begin{equation} \small
\begin{split}
s_{ij} = \exp \big( -\frac{D(RF_i \vert\vert RF_j)}{2\sigma^2} \big),
\end{split}
\label{eq:graph_ij}
\end{equation}
in which $\sigma$ is the parameter controlling the transformation.

Actually,
the similarity graph can be seen as the derivation of a distance graph through the Gaussian kernel.
As the distance graph is built by every pair of RF candidates,
it is a dense one that connects many uncorrelated candidates.
Therefore,
to purify the similarity graph,
we can either keep a fixed number of smallest values in each row/column of the distance graph,
or set a threshold to remove larger values,
leading to the so-called $k$NN graph and $\epsilon$-ball graph~\cite{belkin2003laplacian},
respectively.

It is also worth noting that building the similarity graph is the most costly stage in our computation.
The popular methods usually adopt the dense feature extraction scheme~\cite{lazebnik2006beyond},
which, supposedly $n$  descriptors (with $d$-dimensionality) being extracted in each of the $N$ images from a specific category,
requires computational complexity $O(d n^2 N^2)$ for constructing the graph based on PED.
As the dense extraction of SIFT descriptors consistently leads to thousands of descriptors,
constructing the graph is extremely time-consuming.
To expedite this stage,
we can either turn to  fast approximate $k$NN graph construction~\cite{chen2009fast} or
the original SIFT feature~\cite{lowe2004distinctive},
which incorporates interest point detection and feature descriptor extraction.
But we merely use the original SIFT feature.
Essentially,
with the interest point detection technique in SIFT,
only $n\approx 150$ descriptors are generated in an image of $150\times 150$-pixel resolution.
Then,
it is efficient enough for calculating pairwise PED among RF candidates and constructing the similarity graph in our experiments.
Moreover,
in contrast to the dense extraction scheme that produce most unnecessary descriptors,
such detection technique leads to more meaningful SIFT descriptors in informative regions.


\section{Classifier Design}
\label{sec:classifier}
Our framework is similar to multi-instance learning~\cite{dietterich1997solving},
but is of particularity,
which is especially reflected from principles like center-bias, intra- and inter-image contrast.
By solving the problem of collaborative receptive field learning,
we design a nonparametric classifier by incorporating RF-to-class metric and center-bias penalty.

With the learned RF's in training set,
we put SIFT features of the grids at corresponding positions in a set of pools,
and denote $P_l^c$ to store the descriptors from all training images of the $c^{th}$ class at the specific grid indexed by $l$.
Then,
fed a query image,
our method first extracts SIFT features and $M$ RF candidates (denoted by $\{RF_k\}, k=1,\dots,M $,
and $RF_k = \{X_l^k \vert l=1,\dots,L \}$).
Then,
it predicts the label by comparing RF-to-class distances of all the $C$ categories:
\begin{equation}\small
\begin{split}
c^* = \argmin_c   & \big\{
        \min
        \limits_{%
            \substack{
             k }} %
         \sum_{l=1}^L dist(X^k_l \vert\vert P^c_l) + \lambda_2 q_{k}
        \big \}.
\end{split}
\label{eq:classifier}
\end{equation}
Inspired by~\cite{boiman2008defense},
we exploit KD-tree~\cite{KDtree} to speed up the classification process,
which requires low complexity $O(Nn\log(Nn))$ for training the KD-tree for each category,
and has $O(Cn\log(Nn))$ complexity to predict a query image.

\section{Experimental Validation}
\label{sec:exp}
In this section,
we first qualitatively validate the effectiveness of our method for coRFL over a synthetic dataset in discovering the $RF_+$'s from images.
Then,
we use public benchmarks to quantitatively evaluate our method in object categorization,
including Caltech101~\cite{fei2007learning} and Caltech256~\cite{griffin2007caltech}.
Finally,
we discuss the parameters used in our experiments.

\subsection{Synthetic Data}
\label{ssec:synthetic}

{
\begin{figure*}[t]
\centering	
\includegraphics[width=0.92\textwidth]{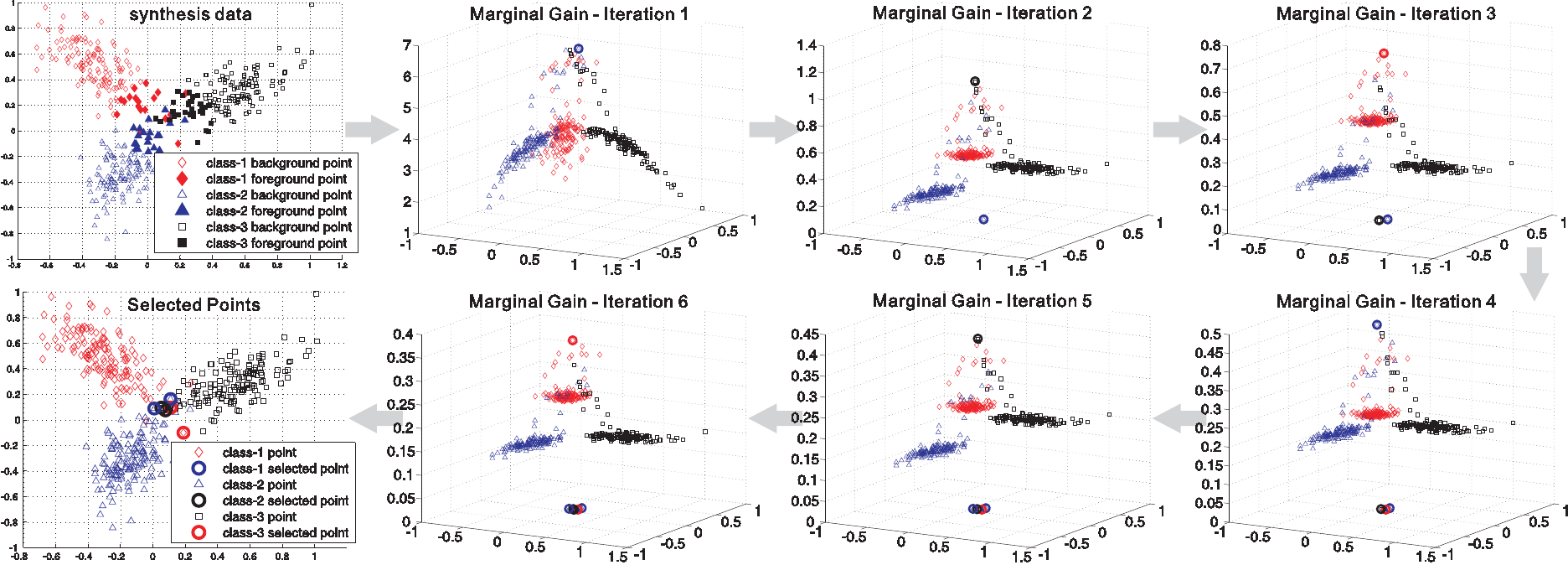}
\caption{ (Best seen in color and zoom in.)
Demonstration of the proposed submodular function in exemplar selection over a synthetic data set.
Leftmost column: display of the generated data points (upper), and the six selected exemplars by the proposed submodular function (down).
The rest columns: marginal gain at each iteration.
This figure illustrates that, with our objective function,
the most similar or correlated exemplars can be found.
}
\label{fig:syntheticDemonstration}
\end{figure*}
}

To generate the dataset,
we use three Gaussian distributions to produce random points in a 2D plate,
as demonstrated by the first upper-left panel in Fig.~\ref{fig:syntheticDemonstration}.
The three clusters can be seen as three images,
and their intersection can be seen as the common objects in the images,
meanwhile,
points far away from the intersection can be imagined as cluttered backgrounds.
Therefore,
our objective function in Eq.~\ref{eq:obj} is expected to find a set of points located in the intersection.
Please note that we set  $\lambda_2 = 0$ as center-bias prior does not apply to the synthetic data;
and  set $\lambda_1 = 2$, $\tau = 2$.
We use the Euclidean distance of their locations in the 2D plate to build the similarity graph with the Gaussian kernel controlled by parameter $\sigma=0.3$.

To better understand the process\footnote{Code is available at Shu Kong's GitHub: https://github.com/aimerykong/coRFL},
we plot the marginal gains at the first six iterations in Fig.~\ref{fig:syntheticDemonstration},
and the six most desirable points in the bottom-left panel.
The optimization is done by the native greedy method.
From the marginal gain at each iteration,
we can see points near the intersection have larger expected gains.
This is owing to our inter-image prior in Eq.~\ref{eq:similarityTerm}.
This figure demonstrates the effectiveness of our approach in finding the most correlated RF's that cover the common foreground object.

\subsection{Benchmark Databases}
\label{ssec:benchmark}

Caltech101 and Caltech256 contain 102 and 256 categories,
and have $9,144$ and $30,607$ images,
respectively.
Caltech256 have higher intra-class variability and higher object location variability (translations and scales) than Caltech101.
We resize every image into no more than $150\times150$-pixel resolution with original aspect ratio.
We follow the common setup on the two benchmarks~\cite{yang2009linear},
\ie 30 images per class are randomly selected as the training set and the rest for testing.
We perform coRFL in each category and set $K=30$ so that exactly 30 $RF_+$'s are extracted for each class.
The average performance after 10 random splits is reported.

\begin{table}[t]
    \centering
    \footnotesize
\caption{Classification accuracies ($\%$) by different methods on the Caltech101 and Caltech256.}
    \begin{tabular}{l r r}
    \toprule
    Method                                                      & Caltech101  & Caltech256 \\
    \midrule
    CDBN~\cite{lee2009convolutional}         & $65.4$                           & -\\
    DN~\cite{zeiler2011adaptive}   & $71.0$                           &  $33.2 \pm 0.8$ \\
    LC-KSVD~\cite{jiang2013label}                & $73.6$                           &  $34.3$ \\
    KSPM~\cite{lazebnik2006beyond}           & $64.6$                           &  $29.5 \pm 0.5$ \\
    ScSPM~\cite{yang2009linear}                  & $73.2$                           &  $34.0 \pm 0.6$ \\
    LLC~\cite{wang2010locality}                    & $73.4$                           &  $41.2$ \\
    RFL~\cite{jia2012beyond}                         & $75.3\pm0.7$               & -       \\
    GMK~\cite{duchenne2011graph}             & $80.3\pm1.2$               & $38.1\pm 0.6$\\
    NBNN~\cite{boiman2008defense}           & $70.4$                           & $37.0$\\
    \midrule
    \textbf{Ours}                                               & $\textbf{83.4} \pm \textbf{1.3}$  & $\textbf{45.7}  \pm \textbf{1.1} $\\
    \bottomrule
    \end{tabular}
  \label{tab:CaltechComparison}
\end{table}

\begin{figure}[t]
        \centering
        \includegraphics[width=.9\linewidth]{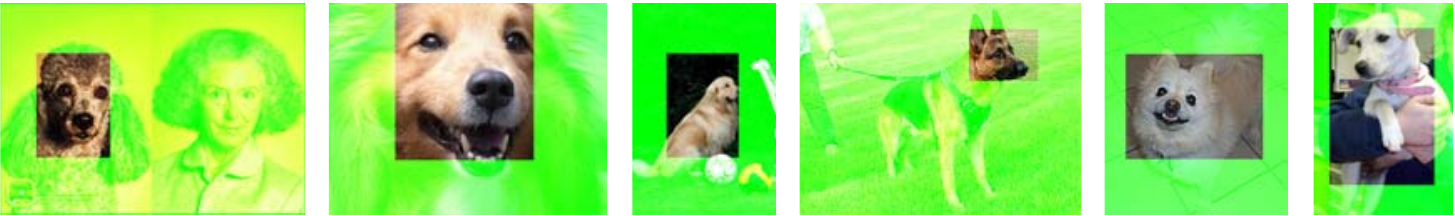}
    \caption{Samples of selected receptive fields over ``dog'' category from Caltech256 (best seen in color and zoom in).}
    \label{fig:dog}
\end{figure}

We compare our method with several state-of-the-art ones.
Most approaches learn mid-level features over low-level ones to represent the overall image,
including
Convolutional Deep Belief Networks (CDBN)~\cite{lee2009convolutional},
adaptive Deconvolutional Networks (DN)~\cite{zeiler2011adaptive},
LC-KSVD~\cite{jiang2013label},
Kernel SPM (KSPM)~\cite{lazebnik2006beyond},
Sparse Coding based SPM (ScSPM)~\cite{yang2009linear},
Locality-constrained Linear Coding (LLC)~\cite{wang2010locality},
Receptive Field Learning (RFL)~\cite{jia2012beyond},
and GMK~\cite{duchenne2011graph}.
CDBN and DN belong to the deep feature learning framework which hierarchically learns adaptive features for image.
Both of them generate mid-level features with the spatial pyramid partition and kernel SVM for classification.
LC-KSVD is a deeper approach that simultaneously learns a linear classifier and a higher-level dictionary over the mid-level features.
The rest methods learn a codebook (consisting of approximate 1024/2048 words for the two benchmarks) over hand-craft  descriptors like SIFT, HoG and Macrofeature~\cite{boureau2010learning};
encode them over the codebook by vector quantization or sparse coding;
and then adopt the pooling technique to obtain the feature vectors for image regions w.r.t a 3-layer-pyramid partition.
Finally,
they concatenate the pooled vectors into a larger one as the image feature and feed into a linear SVM.
Additionally,
the Naive-Bayes Nearest-Neighbor (NBNN)~\cite{boiman2008defense} is closely related to ours,
as it directly uses dense SIFT descriptors, image-to-class metric and NN for classification.
We list the comparisons in Table~\ref{tab:CaltechComparison}.
An illustration of the selected $RF_+$'s by our method are displayed in Fig.~\ref{fig:dog} and the first row in Fig.~\ref{fig:lambda2small}.


As can be seen from Table~\ref{tab:CaltechComparison},
our method outperforms all the others.
It is worth noting the improvement of our method over NBNN attributes to  our PED and SIFT extraction with interest point detection.
Because PED explicitly considers the shape/structure of the objects and interest point detection removes noisy descriptors.
In contrast,
NBNN merely constrains position distances of the dense SIFT to be small,
thus it incorporates noisy descriptors and fails to handle notable changes of object translation and scale.
It is also worth noting that the performance gain brought by our method for Caltech256 is higher than that for Caltech101.
We assume the reason is that the changes of object translation and scale is larger in Caltech256 than those in Caltech101.
Moreover,
we observe that the more cluttered background in the images is,
the better performance of our method achieves in finding the objects.
This is due to the functionality of our objective function that intends to find the most desirable $RF_+$'s,
and leaves behind the assumed $RF_-$'s which hold smaller sum of pairwise similarities.

\subsection{Parameter Discussion}
\label{ssec:parDiscussion}

\begin{figure}[t]
        \centering
        \includegraphics[width=.85\linewidth]{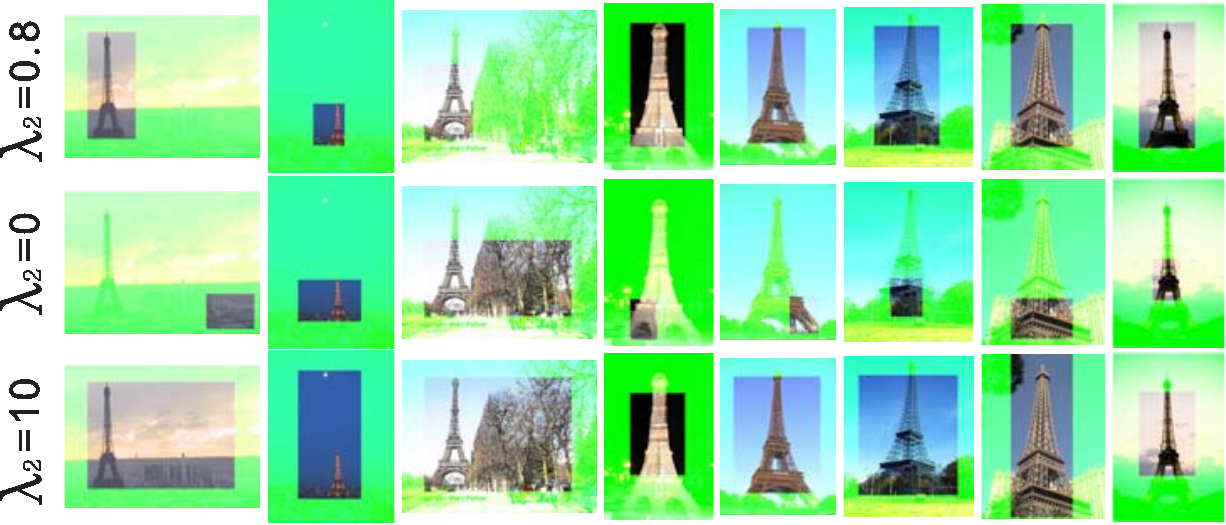}
    \caption{Larger $\lambda_2$ means locating the $RF_+$'s at the center of Caltech256 images with brutal force, while smaller value produces $RF_+$'s that merely capture object parts.
    The real $RF_+$'s can be found with a suitable $\sigma$,
    as the whole object appears in the image center for most images.}
    \label{fig:lambda2small}
\end{figure}

Our work involves several parameters,
including  $\tau$, $\lambda_1$ and $\lambda_2$ in the objective function Eq.~\ref{eq:obj},
and $k$ and $\sigma$ in the Gaussian kernel for constructing the similarity graph.

$\tau$ should be greater than $1$ to ensure the physical meaning of our model.
When varying the value of $\tau$,
we find the classification performance and the visualization of the learned receptive fields do not suffer at all.
The reason we guess is due to our objective function,
which constrains the assumed $RF_-$'s have large PED (small similarities).
Therefore,
larger $\tau$ will indirectly contribute to discovering the $RF_+$'s by finding the most dissimilar $RF_-$'s.
Moreover,
a larger value in $\lambda_1$ guarantees that each training image contribute at least one $RF_+$ (overfitting problem is thus alleviated),
so we merely set $\lambda_1=100$ to ensure this.
$\lambda_2$ controls the mild center-bias constraint,
and has an impact on the performance.
Since most foreground objects appear near the center of images,
a suitable $\lambda_2$ helps to find the real $RF_+$'s.
This can be  demonstrated by Fig.~\ref{fig:lambda2small}.

\begin{figure}[t]
        \centering
        \includegraphics[width=.80\linewidth]{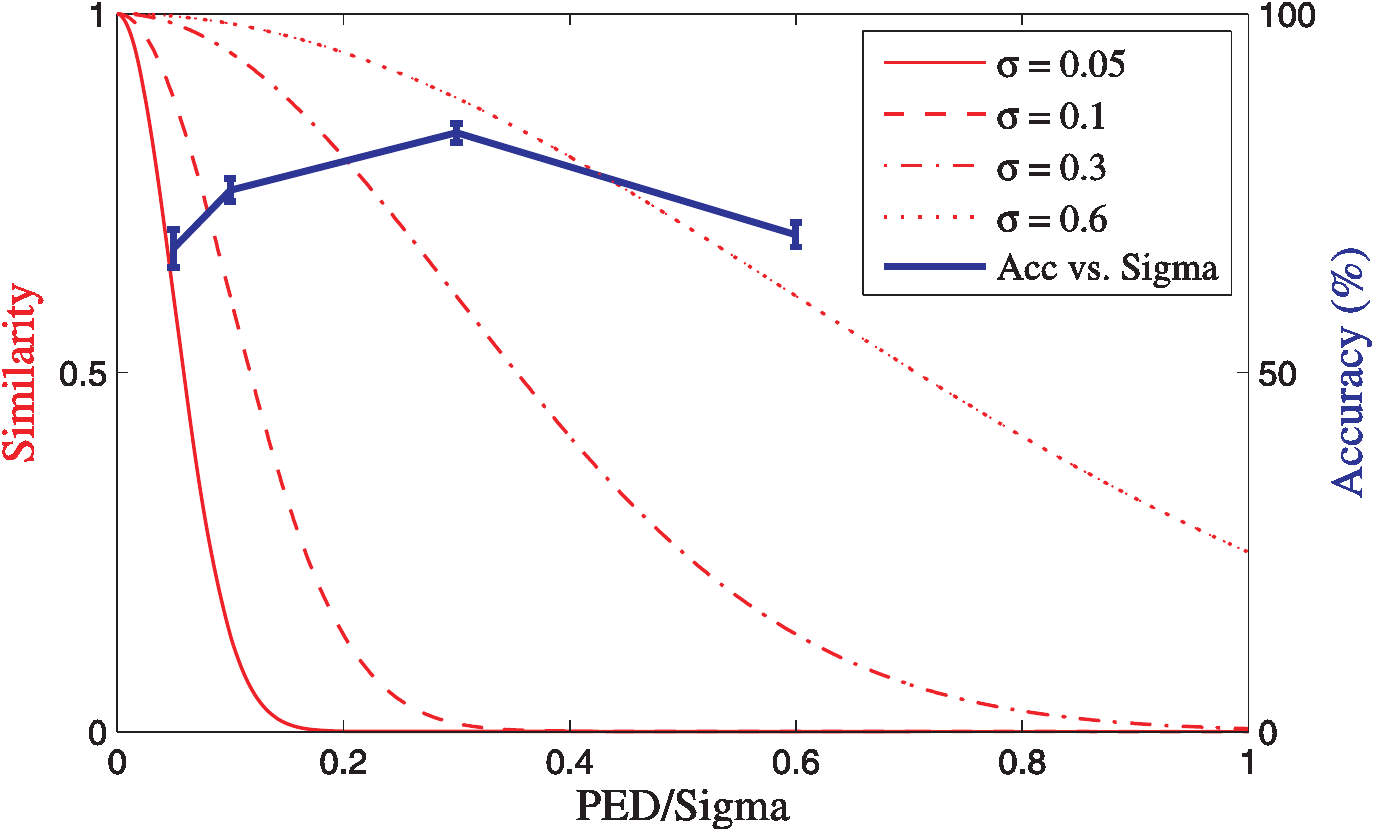}
    \caption{Discussion of $\sigma$ vs. similarity/accuracy in transforming the dissimilarity graph into the similarity graph.}
    \label{fig:paraDiscussion}
\end{figure}

To construct the similarity graph,
we essentially normalize the PED-based distance graph by dividing its largest element,
so that all the entries have values in the range of $[0,1]$.
Then, we transform it into similarity graph with Gaussian kernel controlled by $\sigma$.
We plot the impacts of $\sigma$ in Fig.~\ref{fig:paraDiscussion} over Caltech101 database.
Intuitively,
with the normalized dissimilarity graph,
we can anticipate meaningful outcomes by setting $\sigma \in (0,1)$.
The curve of accuracy vs. $\sigma$ also demonstrates this intuition.
Therefore,
we set $\sigma=0.3$ throughout our work.

\section{Conclusion with Discussion}
\label{sec:conclusion}
In this work,
we introduce a new problem called collaborative receptive field learning (coRFL),
which intends to find receptive fields (RF's) or image regions from multiple images that cover the foreground objects of a common category.
coRFL merely exploits the weak labels without any exact locations of the objects in the image.
By modeling the problem as selecting specific vertices from a similarity graph with consideration of some vision-based priors,
we solve coRFL by a submodular function,
with which a simple greedy algorithm suffices to produce performance-guaranteed solutions.
Furthermore,
we propose the Pyramid Error Distance (PED) to measure pairwise distance of RF's.
We perform coRFL over images of each category to purify the training set,
so that the purified set merely preserves the most meaningful foreground objects.
Moreover,
in consistent with the PED,
we design a simple nonparametric classifier for the final classification.

Our work by no means tries to compete the best results in literature~\cite{DBLP:journals/corr/ZeilerF13},
but it presents several worthwhile research directions.
\begin{itemize}
\item In building the similarity graph,
        we exploit the SIFT feature within the proposed PED.
        Even though fast graph construction techniques can be explored,
        other sophisticated representations for the receptive fields are solicited with the consideration of more robustness and efficiency.
        Especially, learning adaptive features within deep architecture can be exploited to represent the image regions~\cite{KongMidFeaNS}, and learning adaptive metric for matching is also a research direction for image region matching.
\item We model the problem with a simple submodular function by exploiting weak labels,
        but other considerations are worth exploring, \eg semi-supervised learning with a few images providing accurate object locations.
\item Even if our model provides a philosophy of scale and translation invariant region matching,
        we can also consider arbitrary rotation by sophisticated region representations~\cite{KongMVA2013}.
\item Through discussing the parameter $\lambda_2$, we interestingly find that some meaningful patches are selected instead of the whole foreground objects. This motivate us to think about learning the discriminative image patches among images similar to~\cite{singh2012unsupervised, DoerschNIPS13}.
    In particular, our work can also benefit part-based model for specific problems, such as fine-grained recognition~\cite{farrell2011birdlets}.
\item The submodular function also provides a roundabout way for representation learning and instance selection~\cite{krause2010submodular, KongECMLPKDD2013}.
    Modeling problems with proper submodular function is significantly efficient to deal with large-scale data in practice.
\end{itemize}

\bibliography{example_paper}

\begin{thebibliography}{43}
\providecommand{\natexlab}[1]{#1}
\providecommand{\url}[1]{\texttt{#1}}
\expandafter\ifx\csname urlstyle\endcsname\relax
  \providecommand{\doi}[1]{doi: #1}\else
  \providecommand{\doi}{doi: \begingroup \urlstyle{rm}\Url}\fi

\bibitem[Belkin \& Niyogi(2003)Belkin and Niyogi]{belkin2003laplacian}
Belkin, Mikhail and Niyogi, Partha.
\newblock Laplacian eigenmaps for dimensionality reduction and data
  representation.
\newblock \emph{Neural computation}, 15\penalty0 (6):\penalty0 1373--1396,
  2003.

\bibitem[Bentley(1975)]{KDtree}
Bentley, Jon~Louis.
\newblock Multidimensional binary search trees used for associative searching.
\newblock \emph{Commun. ACM}, 18:\penalty0 509--517, 1975.

\bibitem[Boiman et~al.(2008)Boiman, Shechtman, and Irani]{boiman2008defense}
Boiman, Oren, Shechtman, Eli, and Irani, Michal.
\newblock In defense of nearest-neighbor based image classification.
\newblock In \emph{CVPR}, 2008.

\bibitem[Boureau et~al.(2010)Boureau, Bach, LeCun, and
  Ponce]{boureau2010learning}
Boureau, Y-L, Bach, Francis, LeCun, Yann, and Ponce, Jean.
\newblock Learning mid-level features for recognition.
\newblock In \emph{CVPR}, 2010.

\bibitem[Chang et~al.(2011)Chang, Liu, and Lai]{chang2011co}
Chang, Kai-Yueh, Liu, Tyng-Luh, and Lai, Shang-Hong.
\newblock From co-saliency to co-segmentation: An efficient and fully
  unsupervised energy minimization model.
\newblock In \emph{CVPR}, 2011.

\bibitem[Chen et~al.(2009)Chen, Fang, and Saad]{chen2009fast}
Chen, Jie, Fang, Haw-ren, and Saad, Yousef.
\newblock Fast approximate k nn graph construction for high dimensional data
  via recursive lanczos bisection.
\newblock \emph{JMLR}, 10:\penalty0 1989--2012, 2009.

\bibitem[Coates \& Ng(2011)Coates and Ng]{coates2011importance}
Coates, Adam and Ng, Andrew.
\newblock The importance of encoding versus training with sparse coding and
  vector quantization.
\newblock In \emph{ICML}, 2011.

\bibitem[Cormen et~al.(2009)Cormen, Leiserson, Rivest, and
  Stein]{Cormen2001introduction}
Cormen, Thomas~H., Leiserson, Charles~E., Rivest, Ronald~L., and Stein,
  Clifford.
\newblock \emph{Introduction to Algorithms}.
\newblock The MIT Press, 3 edition, 2009.

\bibitem[Dalal \& Triggs(2005)Dalal and Triggs]{dalal2005histograms}
Dalal, Navneet and Triggs, Bill.
\newblock Histograms of oriented gradients for human detection.
\newblock In \emph{CVPR}, 2005.

\bibitem[Dietterich et~al.(1997)Dietterich, Lathrop, and
  Lozano-P{\'e}rez]{dietterich1997solving}
Dietterich, Thomas~G, Lathrop, Richard~H, and Lozano-P{\'e}rez, Tom{\'a}s.
\newblock Solving the multiple instance problem with axis-parallel rectangles.
\newblock \emph{Artificial Intelligence}, 89\penalty0 (1):\penalty0 31--71,
  1997.

\bibitem[Doersch et~al.(2013)Doersch, Gupta, and Efros]{DoerschNIPS13}
Doersch, Carl, Gupta, Abhinav, and Efros, Alexei~A.
\newblock Mid-level visual element discovery as discriminative mode seeking.
\newblock In \emph{NIPS}, 2013.

\bibitem[Domingos(2012)]{DomingosCommACM}
Domingos, Pedro.
\newblock A few useful things to know about machine learning.
\newblock \emph{Commun. ACM}, 55\penalty0 (10):\penalty0 78--87, 2012.

\bibitem[Duchenne et~al.(2011)Duchenne, Joulin, and Ponce]{duchenne2011graph}
Duchenne, Olivier, Joulin, Armand, and Ponce, Jean.
\newblock A graph-matching kernel for object categorization.
\newblock In \emph{ICCV}, 2011.

\bibitem[Farrell et~al.(2011)Farrell, Oza, Zhang, Morariu, Darrell, and
  Davis]{farrell2011birdlets}
Farrell, Ryan, Oza, Om, Zhang, Ning, Morariu, Vlad~I, Darrell, Trevor, and
  Davis, Larry~S.
\newblock Birdlets: Subordinate categorization using volumetric primitives and
  pose-normalized appearance.
\newblock In \emph{ICCV}, 2011.

\bibitem[Fei-Fei et~al.(2007)Fei-Fei, Fergus, and Perona]{fei2007learning}
Fei-Fei, Li, Fergus, Rob, and Perona, Pietro.
\newblock Learning generative visual models from few training examples: An
  incremental bayesian approach tested on 101 object categories.
\newblock \emph{CVIU}, 2007.

\bibitem[Fu et~al.(2013)Fu, Cao, and Tu]{fu2013cluster}
Fu, Huazhu, Cao, Xiaochun, and Tu, Zhuowen.
\newblock Cluster-based co-saliency detection.
\newblock \emph{TIP}, 22\penalty0 (10):\penalty0 3766--3778, 2013.

\bibitem[{Girshick} et~al.(2013){Girshick}, {Donahue}, {Darrell}, and
  {Malik}]{2013arXiv1311.2524G}
{Girshick}, R., {Donahue}, J., {Darrell}, T., and {Malik}, J.
\newblock {Rich feature hierarchies for accurate object detection and semantic
  segmentation}.
\newblock \emph{arXiv}, 2013.

\bibitem[Griffin et~al.(2007)Griffin, Holub, and Perona]{griffin2007caltech}
Griffin, G., Holub, A., and Perona, P.
\newblock {Caltech-256 Object Category Dataset}.
\newblock Technical Report CNS-TR-2007-001, California Institute of Technology,
  2007.

\bibitem[Iyer et~al.(2013)Iyer, Jegelka, and Jeff]{Iyer2013icml}
Iyer, Rishabh, Jegelka, Stefanie, and Jeff, Bilmes.
\newblock fast semidifferential-based submodular function optimization.
\newblock In \emph{ICML}, 2013.

\bibitem[Jia et~al.(2012)Jia, Huang, and Darrell]{jia2012beyond}
Jia, Yangqing, Huang, Chang, and Darrell, Trevor.
\newblock Beyond spatial pyramids: Receptive field learning for pooled image
  features.
\newblock In \emph{CVPR}, 2012.

\bibitem[Jiang et~al.(2013)Jiang, Lin, and Davis]{jiang2013label}
Jiang, Zhuolin, Lin, Zhe, and Davis, Larry~S.
\newblock Label consistent k-svd: Learning a discriminative dictionary for
  recognition.
\newblock \emph{PAMI}, 35\penalty0 (11):\penalty0 2651--2664, 2013.

\bibitem[Kim \& Xing(2013)Kim and Xing]{gunhee-cvpr-13}
Kim, Gunhee and Xing, Eric~P.
\newblock Jointly aligning and segmenting multiple web photo streams for the
  inference of collective photo storylines.
\newblock In \emph{CVPR}, 2013.

\bibitem[Kong \& Wang(2013)Kong and Wang]{KongECMLPKDD2013}
Kong, Shu and Wang, Donghui.
\newblock Learning exemplar-represented manifolds in latent space for
  classification.
\newblock In \emph{ECML-PKDD}. 2013.

\bibitem[Kong et~al.(2014)Kong, Jiang, and Yang]{KongMidFeaNS}
Kong, Shu, Jiang, Zhuolin, and Yang, Qiang.
\newblock Learning mid-level features and modeling neuron selectivity for image
  classification.
\newblock \emph{arXiv preprint arXiv:1401.5535}, 2014.

\bibitem[Krause \& Cevher(2010)Krause and Cevher]{krause2010submodular}
Krause, Andreas and Cevher, Volkan.
\newblock Submodular dictionary selection for sparse representation.
\newblock In \emph{Proceedings of the 27th International Conference on Machine
  Learning (ICML-10)}, pp.\  567--574, 2010.

\bibitem[Krizhevsky et~al.(2012)Krizhevsky, Sutskever, and
  Hinton]{krizhevsky2012imagenet}
Krizhevsky, Alex, Sutskever, Ilya, and Hinton, Geoff.
\newblock Imagenet classification with deep convolutional neural networks.
\newblock In \emph{NIPS}, 2012.

\bibitem[Lazebnik et~al.(2006)Lazebnik, Schmid, and Ponce]{lazebnik2006beyond}
Lazebnik, Svetlana, Schmid, Cordelia, and Ponce, Jean.
\newblock Beyond bags of features: Spatial pyramid matching for recognizing
  natural scene categories.
\newblock In \emph{CVPR}, 2006.

\bibitem[Lee et~al.(2009)Lee, Grosse, Ranganath, and Ng]{lee2009convolutional}
Lee, Honglak, Grosse, Roger, Ranganath, Rajesh, and Ng, Andrew~Y.
\newblock Convolutional deep belief networks for scalable unsupervised learning
  of hierarchical representations.
\newblock In \emph{ICML}, 2009.

\bibitem[Leskovec et~al.(2007)Leskovec, Krause, Guestrin, Faloutsos,
  VanBriesen, and Glance]{leskovec2007cost}
Leskovec, Jure, Krause, Andreas, Guestrin, Carlos, Faloutsos, Christos,
  VanBriesen, Jeanne, and Glance, Natalie.
\newblock Cost-effective outbreak detection in networks.
\newblock In \emph{KDD}, 2007.

\bibitem[Lowe(2004)]{lowe2004distinctive}
Lowe, David~G.
\newblock Distinctive image features from scale-invariant keypoints.
\newblock \emph{IJCV}, 60\penalty0 (2):\penalty0 91--110, 2004.

\bibitem[Nemhauser et~al.(1978)Nemhauser, Wolsey, and
  Fisher]{nemhauser1978analysis}
Nemhauser, George~L, Wolsey, Laurence~A, and Fisher, Marshall~L.
\newblock An analysis of approximations for maximizing submodular set
  functions¡ªi.
\newblock \emph{Mathematical Programming}, 14\penalty0 (1):\penalty0 265--294,
  1978.

\bibitem[Nguyen et~al.(2009)Nguyen, Torresani, de~la Torre, and
  Rother]{nguyen2009weakly}
Nguyen, Minh~Hoai, Torresani, Lorenzo, de~la Torre, Fernando, and Rother,
  Carsten.
\newblock Weakly supervised discriminative localization and classification: a
  joint learning process.
\newblock In \emph{ICCV}, 2009.

\bibitem[Olshausen et~al.(1996)]{olshausen1996emergence}
Olshausen, Bruno~A et~al.
\newblock Emergence of simple-cell receptive field properties by learning a
  sparse code for natural images.
\newblock \emph{Nature}, 1996.

\bibitem[Russakovsky et~al.(2012)Russakovsky, Lin, Yu, and
  Fei-Fei]{russakovsky2012object}
Russakovsky, Olga, Lin, Yuanqing, Yu, Kai, and Fei-Fei, Li.
\newblock Object-centric spatial pooling for image classification.
\newblock In \emph{ECCV}. 2012.

\bibitem[Shi \& Malik(2000)Shi and Malik]{shi2000normalized}
Shi, Jianbo and Malik, Jitendra.
\newblock Normalized cuts and image segmentation.
\newblock \emph{PAMI}, 22\penalty0 (8):\penalty0 888--905, 2000.

\bibitem[Singh et~al.(2012)Singh, Gupta, and Efros]{singh2012unsupervised}
Singh, Saurabh, Gupta, Abhinav, and Efros, Alexei~A.
\newblock Unsupervised discovery of mid-level discriminative patches.
\newblock In \emph{ECCV}. 2012.

\bibitem[Tatler(2007)]{tatler2007central}
Tatler, Benjamin~W.
\newblock The central fixation bias in scene viewing: Selecting an optimal
  viewing position independently of motor biases and image feature
  distributions.
\newblock \emph{Journal of Vision}, 7\penalty0 (14), 2007.

\bibitem[van~de Sande et~al.(2011)van~de Sande, Uijlings, Gevers, and
  Smeulders]{van2011segmentation}
van~de Sande, Koen~EA, Uijlings, Jasper~RR, Gevers, Theo, and Smeulders,
  Arnold~WM.
\newblock Segmentation as selective search for object recognition.
\newblock In \emph{ICCV}, 2011.

\bibitem[Wang \& Kong(2012)Wang and Kong]{KongMVA2013}
Wang, Donghui and Kong, Shu.
\newblock Learning class-specific dictionaries for digit recognition from
  spherical surface of a 3d ball.
\newblock \emph{Machine Vision and Applications}, pp.\  1--15, 2012.

\bibitem[Wang et~al.(2010)Wang, Yang, Yu, Lv, Huang, and
  Gong]{wang2010locality}
Wang, Jinjun, Yang, Jianchao, Yu, Kai, Lv, Fengjun, Huang, Thomas, and Gong,
  Yihong.
\newblock Locality-constrained linear coding for image classification.
\newblock In \emph{CVPR}, 2010.

\bibitem[Yang et~al.(2009)Yang, Yu, Gong, and Huang]{yang2009linear}
Yang, Jianchao, Yu, Kai, Gong, Yihong, and Huang, Thomas.
\newblock Linear spatial pyramid matching using sparse coding for image
  classification.
\newblock In \emph{CVPR}, 2009.

\bibitem[Zeiler \& Fergus(2013)Zeiler and Fergus]{DBLP:journals/corr/ZeilerF13}
Zeiler, Matthew~D. and Fergus, Rob.
\newblock Visualizing and understanding convolutional networks.
\newblock \emph{arXiv}, 2013.

\bibitem[Zeiler et~al.(2011)Zeiler, Taylor, and Fergus]{zeiler2011adaptive}
Zeiler, Matthew~D, Taylor, Graham~W, and Fergus, Rob.
\newblock Adaptive deconvolutional networks for mid and high level feature
  learning.
\newblock In \emph{ICCV}, 2011.

\end{thebibliography}
\bibliographystyle{icml2014}


\section*{Appendix: Proof of Proposition~\ref{proposition:Similarity} and Lemma~\ref{lemma:tau}}
We rewrite the function ${\cal H}(A)$ proposition and lemma as below for presentational convenience:
\begin{equation}\small
{\cal H} (A) = \log \big( \mu + h(\S_{A,A}) - \alpha h(\S_{A,\bar A}) - \beta h(\S_{\bar A, \bar A}) \big).
\end{equation}

\textbf{Proposition: }
There exist proper $\alpha$ and $ \beta$ that make the proposed function ${\cal H}(A)$: $2^{\cal V} \rightarrow \RB$
a monotonically increasing and submodular function.

\textbf{Lemma: }
For any $\tau \in \RB^{*}$,
${\cal H}(A)$  is a monotonically increasing and submodular function
by setting $\alpha =\tau-1$ and $\beta= \tau$.

\textbf{Proof: }
To prove the above proposition and lemma,
we just need to prove the lemma,
as this lemma simply leads to a possible choice of parameter $\alpha$ and $\beta$,
both of which can be tuned by the positive scalar $\tau\in\RB^{*}$.

We  present an auxiliary function as below:
\begin{equation}\small
\begin{split}
{\cal R} (A) = \log \big(\mu + h(\S_{A,{\cal V}})  - \tau h(\S_{\bar A, {\cal V}}) \big).
\end{split}
\end{equation}
With the definition over matrix $\S_{A,B} \in \RB^{\vert A\vert \times \vert B \vert}$:
\begin{equation}\small
h(\S_{A,B}) =
\begin{cases}
\sum_{i\in A} \sum_{j\in B} c_{ij}, & A\not=\varnothing \text{ and }  B\not=\varnothing, \\
0, & otherwise.
\end{cases}
\end{equation}
we can easily derive that:
\begin{equation}\small
\begin{split}
{\cal R} (A)
    = & \log \big( \mu+ h(\S_{A,{\cal V}})  - \tau h(\S_{\bar A, {\cal V}}) \big) \\
    = & \log \big( \mu + h(\S_{A,A}) + h(\S_{A,\bar A}) - \tau h(\S_{\bar A, A})  - \tau h(\S_{\bar A, \bar A}) \big) \\
    = & \log \big( \mu + h(\S_{A,A}) - (\tau -1 )h(\S_{A,\bar A}) - \tau h(\S_{\bar A, \bar A}) \big).
\end{split}
\end{equation}
Therefore,
with this auxiliary function,
we have $\alpha=\tau - 1$ and $\beta=\tau$ that satisfy ${\cal H} (A)$.
Now,
to prove the proposition and the lemma,
we can turn to show ${\cal G} (A) $ is a monotonically increasing and submodular function.

To this end,
we split the parts in ${\cal R} (A)$ as below:
\begin{equation}\footnotesize
\begin{split}
{\cal R} (A)  = &  \log \big( \mu + h(\S_{A,{\cal V}})  - \tau h(\S_{\bar A, {\cal V}}) \big) \\
= &  \log \big( \mu +
\sum\limits_{%
            \substack{
            i\in A \\
            j\in {\cal V} }%
            }
            s_{ij} - \tau \sum\limits_{%
            \substack{
            i\in \bar A \\
            j\in {\cal V} }%
            }  s_{ij} \big) \\
= &  \log \big( \mu \mu +  \sum\limits_{%
            \substack{
            i\in A \\
            j\in {\cal V} }%
            }  s_{ij}
        - \tau \sum\limits_{%
            \substack{
            i\in \bar A \\
            j\in {\cal V} }%
            }   s_{ij}
        - \tau \sum\limits_{%
            \substack{
            i\in A \\
            j\in {\cal V} }%
            }  s_{ij} + \tau \sum\limits_{%
            \substack{
            i\in A \\
            j\in {\cal V} }%
            }  s_{ij} \big) \\
= &  \log \big( \mu + (\tau + 1) \sum\limits_{%
            \substack{
            i\in A \\
            j\in {\cal V} }%
            } s_{ij}
        - \tau \sum\limits_{%
            \substack{
            i\in {\cal V} \\
            j\in {\cal V} }%
            } s_{ij} \big) \\
= &  \log \big(\mu - \tau h(\S_{{\cal V},{\cal V}})   + (\tau + 1) h(\S_{A, {\cal V}}) \big).
\end{split}
\end{equation}

\textbf{monotonically increasing: }
By definition, if $A = \varnothing$, then we have ${\cal R}(A)   \ge 0$;
moreover, by denoting $\Delta = \mu - \tau h(\S_{{\cal V},{\cal V}})   + (\tau + 1) h(\S_{A, {\cal V}})$,
we have:
\begin{equation}\footnotesize
\begin{split}
& {\cal R}(A\cup a) - {\cal R}(A) \\
= & \log \frac{  \mu - \tau h(\S_{{\cal V},{\cal V}})   + (\tau + 1) h(\S_{A \cup a, {\cal V}}) }
    { \mu - \tau h(\S_{{\cal V},{\cal V}})   + (\tau + 1) h(\S_{A, {\cal V}}) } \\
= & \log \frac{  \mu - \tau h(\S_{{\cal V},{\cal V}})   + (\tau + 1) \big( h({\S_{A, {\cal V}}}) + \sum_{j} s_{aj}  \big)             }
    { \mu - \tau h(\S_{{\cal V},{\cal V}})   + (\tau + 1) h(\S_{A, {\cal V}}) } \\
= & \log \frac{  \Delta  + (\tau + 1)   \sum_{j} s_{aj}            }
    { \Delta } \\
= & \log \big( 1+ \frac{  (\tau + 1)   \sum_{j} s_{aj}            }
    { \Delta } \big) \\
\ge & 0.
 \end{split}
\end{equation}
Therefore,
${\cal R}(A)$ is monotonically increasing.

\textbf{Submodulary: }
Previous derivation leads to the following:
\begin{equation}\footnotesize
\begin{split}
& {\cal R}(A\cup a) - {\cal R}(A) \\
= & \log \frac{  \mu - \tau h(\S_{{\cal V},{\cal V}})   + (\tau + 1) h(\S_{A \cup a, {\cal V}}) }
    { \mu - \tau h(\S_{{\cal V},{\cal V}})   + (\tau + 1) h(\S_{A, {\cal V}}) } \\
= & \log \frac{  \mu - \tau h(\S_{{\cal V},{\cal V}})   + (\tau + 1) \big( h({\S_{A, {\cal V}}}) + \sum_{j} s_{aj}  \big)             }
    { \mu - \tau h(\S_{{\cal V},{\cal V}})   + (\tau + 1) h(\S_{A, {\cal V}}) } \\
= & \log \frac{  \Delta  + (\tau + 1)   \sum_{j} s_{aj}            }
    { \Delta }. \\
 \end{split}
\end{equation}
Similarly,
we derive:
\begin{equation}
\small
\begin{split}
& {\cal R}(A\cup a \cup b) - {\cal R}(A \cup b) \\
= & \log \frac{  \mu - \tau h(\S_{{\cal V},{\cal V}})   + (\tau + 1) h(\S_{A \cup a \cup b, {\cal V}}) }
    { \mu - \tau h(\S_{{\cal V},{\cal V}})   + (\tau + 1) h(\S_{A \cup b, {\cal V}}) } \\
= & \log \frac{  \mu - \tau h(\S_{{\cal V},{\cal V}})   + (\tau + 1) \big( h({\S_{A, {\cal V}}}) + \sum_{j} s_{aj} + \sum_{j} s_{bj} \big)             }
    { \mu - \tau h(\S_{{\cal V},{\cal V}})   + (\tau + 1) \big( h(\S_{A, {\cal V}}) + \sum_{j} s_{bj} \big)  } \\
= & \log \frac{ \Delta + (\tau + 1) (\sum_{j} s_{aj} + \sum_{j} s_{bj})             }
    { \Delta  + (\tau + 1)  \sum_{j} s_{bj} } \\
 \end{split}
\end{equation}
where $\Delta =\mu - \tau h(\S_{{\cal V},{\cal V}}) + (\tau + 1) h(\S_{A, {\cal V}}) $.
Let $\mu = 1 + \tau h(\S_{{\cal V}, {\cal V}})$,
and denote $x_1 = \Delta$ and $x_2 =  \Delta  + (\tau + 1)  \sum_{j} s_{bj}$,
we have $0 \le x_1 \le x_2$,
and thus:
\begin{equation}\small
\begin{split}
&  x_1 \le x_2 \\
\Longleftrightarrow &  \frac {1} {x_1} \ge \frac{1}{x_2} \\
\Longleftrightarrow &  \frac {(\tau + 1) \sum_{j} s_{aj} } {x_1} \ge \frac{ (\tau + 1) \sum_{j} s_{aj}  }{x_2} \\
\Longleftrightarrow &  1+\frac {(\tau + 1) \sum_{j} s_{aj} } {x_1} \ge 1+\frac{ (\tau + 1) \sum_{j} s_{aj}  }{x_2} \\
\Longleftrightarrow &  \frac {x_1 + (\tau + 1) \sum_{j} s_{aj} } {x_1} \ge \frac{x_2 + (\tau + 1) \sum_{j} s_{aj}  }{x_2} \\
\Longleftrightarrow &  \log \frac {x_1 + (\tau + 1) \sum_{j} s_{aj} } {x_1} \ge \log \frac{x_2 + (\tau + 1) \sum_{j} s_{aj}  }{x_2} \\
\Longleftrightarrow & \big( {\cal G}(A\cup a) - {\cal G}(A) \big) \ge \big( {\cal G}(A\cup a \cup b) - {\cal G}(A \cup b) \big). \\
\end{split}
\end{equation}
Therefore,
the auxiliary function ${\cal R}(A)$ is submodular.

\textbf{Summary: }
Since the auxiliary ${\cal R} (A)$ is a monotonically increasing and submodular function,
with the relationship between $\tau$ and $\alpha=\tau-1$ and $\beta=\tau$,
we show the derived ${\cal H}_{\tau}{ (A)}$ is a monotonically increasing and submodular function.

End of proof. $\blacksquare$

\section*{Appendix: Proof of Proposition~\ref{proposition:property_balance} }
\textbf{Proposition: }
With the function defined as below over vector $\x=[x_1,\dots,x_N] \succcurlyeq 0$):
\begin{equation} \small
\begin{split}
g(\x) = & \sum_{j=1}^N \log( x_j + 1), \\
\end{split}
\nonumber
\end{equation}
if $x_c \le x_i, \forall i \not = c$,
then $g(x_c+1 \vert \x) \ge  g(x_i+1 \vert \x)$,
where $g(x_i+1 \vert \x)$ is defined as:
$g(x_i+1 \vert \x) = \sum_{ j\not=i}  \log( x_j + 1) + \log( x_i + 1 + 1).$

\textbf{Proof: }
With $0 \le x_c \le x_i$, $\forall i\not= c$,
we writing down $g(x_c+1 \vert \x)$ and $g(x_i+1 \vert \x)$ as below:
\begin{equation}\tiny
\begin{split}
g(x_c+1 \vert \x) = & \sum_{j\not= c, j\not=i} \log(x_j+1) + \log(x_c+1+1) + \log(x_i+1), \\
g(x_i+1 \vert \x) = & \sum_{j\not= c, j\not=i} \log(x_j+1) + \log(x_c+1) + \log(x_i+1+1).  \\
\end{split}
\end{equation}
Then, we have:
\begin{equation}\tiny
\begin{split}
    & g(x_c+1 \vert \x) - g(x_i+1 \vert \x)\\
= & \log(x_c+1+1)+\log(x_i+1)- \log(x_c+1)-\log(x_i+1+1) \\
= & \log(x_c+1+1)(x_i+1)- \log(x_c+1)(x_i+1+1) \\
= &  \log \frac{ (x_c+1+1)(x_i+1)}{(x_c+1)(x_i+1+1)}  \\
= &  \log \frac{ (x_c+1)(x_i+1) + (x_i+1)}{(x_c+1)(x_i+1) + (x_c+1)}.  \\
\end{split}
\end{equation}
As $0 \le x_c \le x_i, \forall i\not= c$,
we have:
\begin{equation}\tiny
\begin{split}
    &  x_c \le x_i \\
\Longleftrightarrow & (x_c+1)(x_i+1) +  1+ x_c \le(x_c+1)(x_i+1) +  1+ x_i  \\
\Longleftrightarrow &  \frac{ (x_c+1)(x_i+1) + (x_i+1)}{(x_c+1)(x_i+1) + (x_c+1)} \ge 1 \\
\Longleftrightarrow &  \log \frac{ (x_c+1)(x_i+1) + (x_i+1)}{(x_c+1)(x_i+1) + (x_c+1)} \ge 0 \\
\Longleftrightarrow &  g(x_c+1 \vert \x) - g(x_i+1 \vert \x) \ge 0.
\end{split}
\end{equation}
End of proof. $\blacksquare$

\section*{Appendix: Proof of Proposition~\ref{proposition:balance}}
The proposed function as below to maximize is a monotonically increasing and submodular function:
\begin{equation}
            \begin{split}
            {\cal G}(A) = & \sum_{j=1}^N \log( \vert A_{j}\vert + 1), \\
            \text{s.t. } &  \cup_{j=1}^{N} A_j = A,  A_i\cap A_j = \varnothing, \forall i\not=j.
            \end{split}
\end{equation}

\textbf{Proof: }
To prove ${\cal G}(A)$ is monotonically increasing,
we just need to show ${\cal G}(A\cup a) - {\cal G}(A) \ge 0$, where $a \in {\cal V}$ and $a \not\in A$.
Without of lose of generality,
we can suppose $a$ comes from the $i^{th}$ image (note $a \not\in A_i$),
therefore $a$ is added to $A_i$.
With simple derivations,
we have:
\begin{equation}
\begin{split}
    {\cal G}(A\cup a) - {\cal G}(A) = & \log( \vert A_{i} + a\vert + 1) - \log( \vert A_{i}\vert + 1)
\end{split}
\end{equation}
Here $\vert\cdot\vert$ means the cardinality.
we can denote $x = \vert A_{i}\vert + 1$ is a positive integer,
hence it is easy to see:
\begin{equation}
\begin{split}
{\cal G}(A\cup a) - {\cal G}(A) = \log( x + 1) -\log( x )  > 0.
\end{split}
\end{equation}
Therefore,
${\cal G}(A)$ is a strictly monotonically increasing function.

Moreover, for its submodularity,
we need to show the following for $a\not \in A$ and $b \not\in A$ ($a\not=b$, otherwise equality is achieved):
\begin{equation}
\begin{split}
    {\cal G}(A\cup a) - {\cal G}(A) \ge {\cal G}(A\cup b \cup  a) - {\cal G}(A \cup b).
\end{split}
\end{equation}
There are two cases,
$a$ and $b$ come from  a common image, or two different ones.

For the first case,
suppose $a$ and $b$ come from the $i^{th}$ image,
then we have:
\begin{equation} \footnotesize
\begin{split}
        & \big({\cal G}(A\cup a) - {\cal G}(A)\big)  - \big({\cal G}(A\cup b \cup  a) - {\cal G}(A \cup b) \big) \\
    = & {\cal G}(A\cup a) - {\cal G}(A)  - {\cal G}(A\cup b \cup  a) + {\cal G}(A \cup b)  \\
    = & \sum_{j\not= i}^N \log( \vert A_{j}\vert + 1) + \log(\vert A_i \cup a \vert + 1) \\
        & - \sum_{j\not= i}^N \log( \vert A_{j}\vert + 1) - \log(\vert A_i \vert + 1) \\
        & - \sum_{j\not= i}^N \log( \vert A_{j}\vert + 1) - \log(\vert A_i \cup a \cup b \vert + 1) \\
        & + \sum_{j\not= i}^N \log( \vert A_{j}\vert + 1) + \log(\vert A_i \cup b \vert + 1) \\
    = &  \log(\vert A_i \cup a \vert + 1) - \log(\vert A_i \vert + 1) \\
        & - \log(\vert A_i \cup a \cup b \vert + 1)  + \log(\vert A_i \cup b \vert + 1) \\
    = & \big( \log(x + 1) - \log(x) \big) -  \big(\log(x + 2) - \log(x + 1) \big), \\
\end{split}
\end{equation}
where $x$ is a positive integer.
Now, the question turns to proving $\log(x + 1)$ is a concave function.
This is obvious,
and thus proof done.

For the second case,
\ie $a$ and $b$ come from different images,
we assume $a$ and $b$ come from the $i^{th}$ and the $k^{th}$ image, respectively.
Then we have:
\begin{equation} \tiny
\begin{split}
        & \big({\cal G}(A\cup a) - {\cal G}(A)\big)  - \big({\cal G}(A\cup b \cup  a) - {\cal G}(A \cup b) \big) \\
    = & {\cal G}(A\cup a) - {\cal G}(A)  - {\cal G}(A\cup b \cup  a) + {\cal G}(A \cup b)  \\
    = & \sum_{j\not= i}^N \log( \vert A_{j}\vert + 1) + \log(\vert A_i \cup a \vert + 1) \\
        & - \sum_{j\not= i}^N \log( \vert A_{j}\vert + 1) - \log(\vert A_i \vert + 1) \\
        & - \sum_{j\not= i, j\not = k}^N \log( \vert A_{j}\vert + 1) - \log(\vert A_i \cup a \vert + 1) - \log(\vert A_k \cup b \vert + 1) \\
        & + \sum_{j\not= i, j\not = k}^N \log( \vert A_{j}\vert + 1) + \log(\vert A_i  \vert + 1) + \log(\vert A_k \cup b \vert + 1) \\
    = &  \log(\vert A_i \cup a \vert + 1) - \log(\vert A_i \vert + 1) \\
        & - \log(\vert A_i \cup a \vert + 1)  + \log(\vert A_i \vert + 1) \\
    = & 0 \\
\end{split}
\end{equation}
Therefore,
in this case,
${\cal G}(A)$ is a modular function.

Overall,
we prove the function ${\cal G}(A)$ is a monotonically increasing and submodular function.

End of proof.$\blacksquare$

\section*{Appendix: Proof of Proposition~\ref{proposition:obj} }
We rewrite the function ${\cal F}(A)$ and the  proposition as below for presentational convenience:
\begin{equation}
\begin{split}
{\cal F}(A) \equiv {\cal H}_{\tau} (A) +  \lambda_1 {\cal G} (A) + \lambda_2 \Vert \q_{(A)}\Vert_1.
\end{split}
\end{equation}

\textbf{Proposition: }
The proposed function ${\cal F}(A) $ is a monotonically increasing and submodular function,
and induces a matroid ${\cal M} = ({\cal V}, {\cal I})$,
where ${\cal V}$ is the point set, and $\cal I$ is the collection of subsets $A \subseteq {\cal V}$.

\textbf{Proof: }
As we previously show ${\cal H}_{\tau} (A)$ and $g(A)$ are monotonically increasing and submodular functions,
now we just need to fucus on $\Vert \q_{A}\Vert_1$.
By definition,
$q_i \ge 0, \forall i$,
and
\begin{equation}
\begin{split}
\Vert \q_{A}\Vert_1 = \sum_{i \in A}q_i.
\end{split}
\end{equation}
It is easy to see:
\begin{equation}
\begin{split}
\Vert \q_{A\cup a}\Vert_1 - \Vert \q_{A}\Vert_1 & = q_a  \ge 0,
\end{split}
\end{equation}
thus $\Vert \q_{A}\Vert_1$ is monotonically increasing.

Furthermore,
we have:
\begin{equation}
\begin{split}
& \big(\Vert \q_{A\cup a }\Vert_1 - \Vert \q_{A }\Vert_1\big) -
\big(
\Vert \q_{A\cup a\cup b}\Vert_1 - \Vert \q_{A\cup b}\Vert_1 \big)\\
= & q_a - q_a \\
= & 0.
\end{split}
\end{equation}
Therefore,
$\Vert \q_A\Vert_1$ is a modular function.

In sum,
$\Vert \q_A\Vert_1$ is a monotonically increasing and modular function;
and thus,
the objective function is  a monotonically increasing and submodular function.

\textbf{matroid: }
The proposed objective function induces a matroid ${\cal M} = ({\cal V}, {\cal I})$,
where ${\cal V}$ is the ground set, and $\cal I$ is a family of feasible solution sets.

Proof focuses on the following three conditions:
\begin{enumerate}
  \item $\varnothing \in {\cal I} $: the function start with $\varnothing$ as defined.
  \item (Hereditary property): If $A \subseteq B$ and $B \in \cal I$, then $A \in \cal I$;
  \item (Exchange property): If $A \in \cal I$, $B \in \cal I$ and $\vert A \vert < \vert B \vert$,
            there is an element $e \in B-A$ such that $A\cup e \in \cal I$.
\end{enumerate}
As there is no constraint on the matroid posed in ${\cal F}(A)$,
our objective function induces the desired set $A$ from a uniform matroid.

End of proof. $\blacksquare$

\section*{Appendix: Other Details in Implementation}
A trick to build the dissimilarity graph is to correlate each pair of images with their RF's by selecting fixed number (say 3) of nearest RF's with brutal force,
then we smooth the graph by keeping the fixed number (say the number of training images of each category) of entries with the smallest values and derive the $k$NN graph.

\section*{Appendix: More Results on the Synthetic Data}

\begin{figure*}[t]
\begin{center}
   \includegraphics[width=0.960\linewidth]{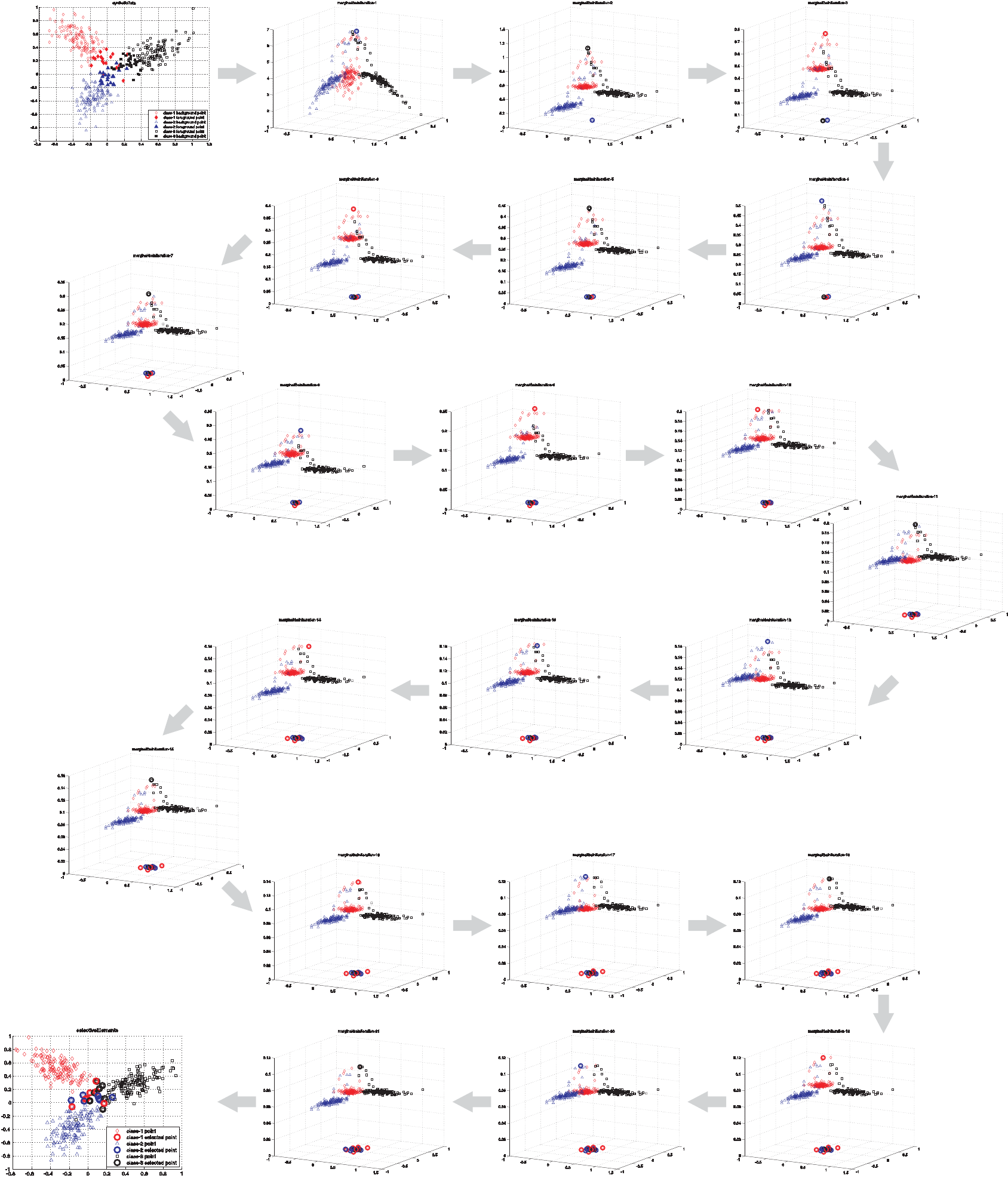}
\end{center}
   \caption{ More iterations over the synthetic dataset to select more receptive fields.
   It can be seen that all the selected data lie on the intersection of the three classes/clusters.
   This demonstrates the effectiveness of the proposed method.}
\label{fig:SyntheticDemonstration_more}
\end{figure*}

Fig.~\ref{fig:SyntheticDemonstration_more} presents more illustrations on the marginal gains of different iterations.
We can see after sufficient iterations,
all the $RF_+$'s that are assumed to be most correlated are found finally.

\section*{Appendix: More Results of Caltech256}

\begin{figure*}[t]
\centering	
\includegraphics[width=0.90\textwidth]{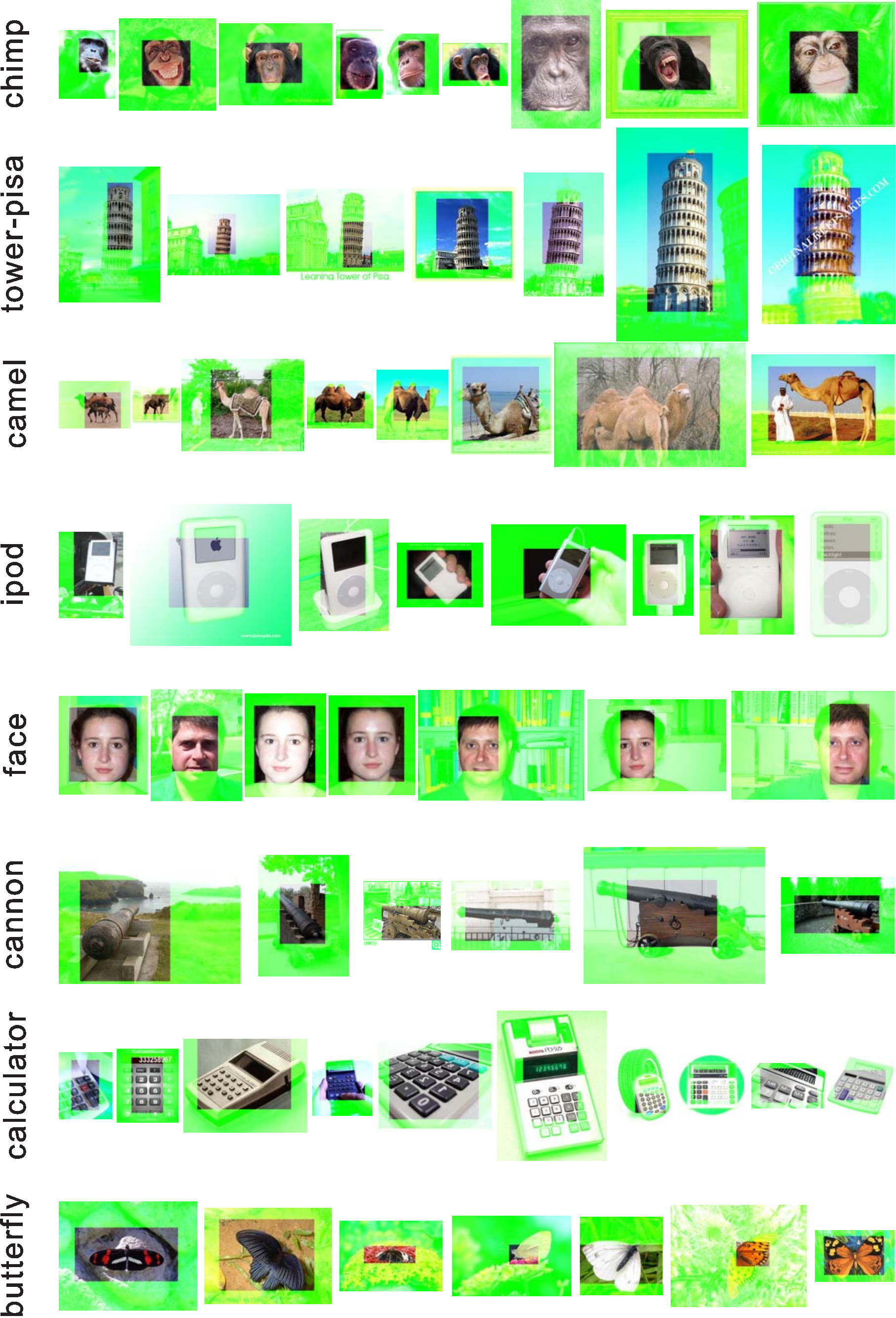}
\caption{ (Best seen in color.) Further illustration of the learned receptive fields over images from Caltech256. }
\label{fig:moreCaltech256}
\end{figure*}

Fig.~\ref{fig:moreCaltech256} displays more results of the learned receptive fields over images from Caltech256.
From the figure,
we can see the most informative RF's are found in the images.
This demonstrates the effectiveness of our method.

\end{document}